\documentclass[letterpaper, 10 pt, conference]{ieeeconf}  

\IEEEoverridecommandlockouts                              

\usepackage{graphics} 
\usepackage{epsfig} 
\usepackage{mathptmx} 
\usepackage{times} 
\usepackage{amsmath} 
\usepackage{amssymb}  
\usepackage{hyperref}
\usepackage{type1cm}

\title{\LARGE \bf
Efficient Disruption of Criminal Networks through Multi-Objective Genetic Algorithms
}

\author{Yehezkiel Darmadi$^{1}$, Thanh Thi Nguyen$^{2*}$ and Campbell Wilson$^{2}$
\thanks{$^{1}$Faculty of Information Technology, Monash University, Melbourne, Victoria, 3800, Australia
        {\tt\small ydar0001@student.monash.edu}}%
\thanks{$^{2}$AiLECS Lab, Faculty of Information Technology, Monash University, Melbourne, Victoria, 3800, Australia
        {\tt\small \{thanh.nguyen9,campbell.wilson\}@monash.edu}}%
\thanks{$^{*}$Corresponding author.}
}

\begin{document}

\maketitle
\thispagestyle{empty}
\pagestyle{empty}

\begin{abstract}

Criminal networks, such as the Sicilian Mafia, pose substantial threats to public safety, national security, and economic stability. Outdated  disruption methods with a focus on removing influential individuals or key players have proven ineffective due to the covertness of the network. Thus, researchers have been trying to apply Social Network Analysis (SNA) techniques, such as centrality-based measures, to identify key players. However, removing individuals with high centrality often proves to be inefficient, as it does not mimic the real-world scenarios that Law Enforcement Agencies (LEAs) face. For instance, the operational costs limit the LEAs from exploiting the results of the centrality-based methods. This study proposes a multi-objective optimisation framework like the Weighted Sum Genetic Algorithm (WS-GA) and the Non-dominated Sorting Genetic Algorithm II (NSGA-II) to identify disruption strategies that balance two conflicting goals, maximising fragmentation and minimising operational cost which is captured by the spatial distance between nodes and the nearest LEA headquarters. The study utilises the ``Montagna Operation'' dataset for the experiments. The results demonstrate that although centrality-based approaches can fragment network effectively, they tend to incur higher operational costs. In contrast, the proposed algorithms achieve comparable disruption outcomes with significantly lower operational costs. The contribution of this work lies in incorporating operational costs in a form of spatial distance constraints into disruption strategy, which has been largely overlooked in prior studies. This research offers a scalable multi-objective capability that improves practical application of SNA in guiding LEAs in disrupting criminal networks more efficiently and strategically.

\end{abstract}

\section{INTRODUCTION}

Criminal networks such as drug trafficking cartels, terrorist groups, and human trafficking organisations, pose substantial risks to public safety, national security, and economic stability. These covert organisations, such as the Sicilian Mafia, have taken control by influencing economic, social and political sectors of entire countries \cite{ficara2021multilayer}. Given such threats, Law Enforcement Agencies (LEAs) need to dismantle these organisations before they become highly threatening. 

Due to the covert and resilient nature of criminal networks, LEAs often find it difficult to dismantle them, which are frequently structured to remain operational despite the removal of key members \cite{mcandrew2021structural}. Thus, the challenge lies in targeting the right individuals to be removed. By targeting key players, LEAs can disrupt the flow of information, resources, and control within the organisation. Traditional methods, such as targeting high-ranking individual proved to be insufficient because these networks are designed to adapt quickly when such individuals are removed \cite{krebs2002uncloaking}. This has led to a shift towards more quantitative techniques that could map the relationships within the networks, giving more strategic insight and interference.

In recent years, Social Network Analysis (SNA) has been utilised to analyse criminal networks. SNA provides insight into the criminal network's structure by identifying the key players derived from metrics of centrality such as degree and betweenness centrality. These metrics rank individuals based on their importance, influence, or position within the network. Targeting these individuals will disrupt the network's ability to communicate and operate \cite{xu2005criminal}. 

Nevertheless, SNA often causes inefficiencies in disruption strategies \cite{ficara2023human}. By focusing excessively on high-centrality individuals to maximise network disruption, these approaches may lead to unnecessary use of time and resources. Because, peripheral actors may play important roles due to decentralised nature of criminal networks \cite{bright2015disrupting}. Moreover, LEAs often have limited operational resources and thus must be carefully considered when disrupting criminal networks~\cite{basu2021identifying}. Thus, LEAs must adopt more efficient strategies that optimise both impact and operational cost. 

Earlier research has overlooked the importance of optimising the selection of key players removed in order to efficiently manage resource allocation while achieving maximum network disruption. Thus, a clear approach is needed to optimise the operational cost, but also has the maximum network disruption. For example, removing criminals that are far away from the LEA's operational ground would be more time consuming and expensive compared to targeting those who are geographically closer. This is important to LEAs, because they have limited resources (time, budget, and personnel). This highlights the need for selecting targets who are both closer and impactful.

To address this gap, this study applied multi-objective genetic algorithms, Weighted Sum Genetic Algorithm (WA-SG) and Non-dominated Sorting Genetic Algorithm II (NSGA-II), to identify a set of selected nodes that maximises network disruption while minimising the operational costs. The objectives are described by: the normalised largest connected component ($\rho$), measuring network fragmentation, and spatial distance ($D$), representing the operational cost. This scalable approach supports law enforcement by identifying key players for removal while balancing disruption impact and operational constraint.

\section{Related Work}

In criminal networks, disruption refers to efforts aimed at hindering or completely stopping their operations. This translates to reducing the flow of information, reducing the capacity of individuals to perform their role, or significantly slowing decision-making processes within the network \cite{krebs2002uncloaking, xu2005criminal}, thereby impacting its functionality. Ultimately, effective disruption should minimise the harm caused by the these networks by weakening their ability to engage in criminal activities.

One approach to simulate the disruption is through node removal strategies and analysing their impact on the network. These strategies can involve sequential or block removal, reflecting real-world LEA operations. The removal of nodes are typically based on criteria such as centrality measures or specific roles \cite{cavallaro2020disrupting, duxbury2020responsiveness, jia2024network}.

There are multiple methods for measuring disruption effectiveness in a network, such as the size of the largest connected components, which indicates network fragmentation, and average geodesic distance, which quantifies the overall connectivity~\cite{cavallaro2020disrupting,ficara2022covert}. These metrics reflect different aspects of network disruption objectives.

Despite the wide range of disruption metrics, each aligned with different objectives, selecting context-appropriate measurements is essential for guiding LEA disruption strategies. The chosen metric shapes how disruption is defined, measured, and addressed \cite{duxbury2019criminal}. This enables systematic evaluation of the effectiveness of disruption strategies, allowing LEAs to refine their approaches and improve efficiency. Furthermore, it helps identify factors contributing to both the vulnerability and resilience of criminal networks, thus supporting the development of more targeted disruption strategies \cite{cavallaro2020disrupting}. Finally, it provides a structured foundation for justifying, guiding, and monitoring the use of resources allocated to disruption-focused efforts.

\section{Methodology}

\subsection{Dataset: Sicilian Mafia ``Montagna Operation''}

This study uses the Sicilian Mafia dataset, specifically on the ``Montagna Operation'', to conduct simulations. The dataset is publicly available on \href{https://zenodo.org/records/3938818}{Zenodo} \cite{cavallaro2020disrupting} and is based on judicial documentation and investigation records between 2003 and 2007. The operation targeted the ``Mistretta'' and ``Batanesi'' families, who infiltrated various economic activities, including public works, through a cartel of entrepreneurs intimately affiliated with the mafia in northeastern Sicily \cite{ficara2021multilayer}.

The dataset was anonymised by the original authors to mitigate privacy and confidentiality converns, ensuring ethical compliance \cite{ficara2021multilayer,jia2024network}. Due to the covert nature of the criminal network, acquiring such data is very rare, making this dataset particularly valuable for research on network structures and disruption strategies. However, it is inherently biased toward known criminal actors. In reality, many individuals involved in criminal activities remain unidentified and thus absent from the dataset, leading to potential incompleteness.

\begin{figure}[ht]
    \centering
    \begin{minipage}{0.45\textwidth}
        \centering
        \includegraphics[width=\textwidth]{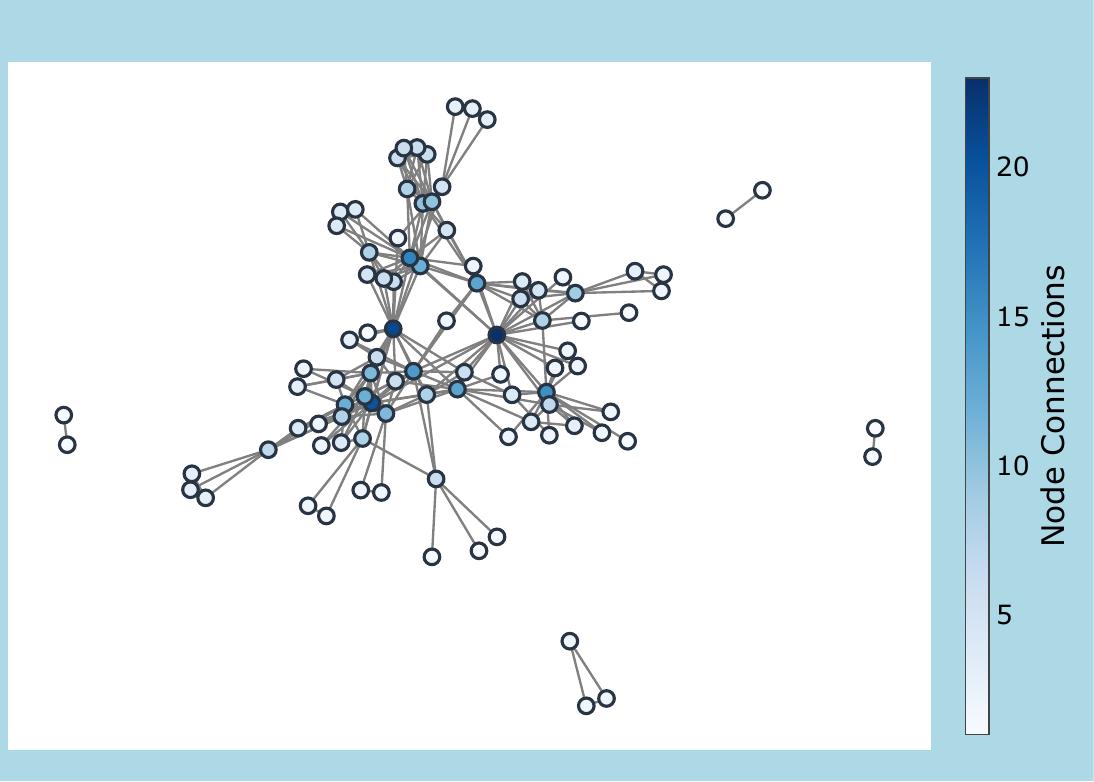} 
        \caption{An example network derived from the \textit{physical meeting} data subset in the Sicilian Mafia ``Montagna Operation'' dataset, featuring 95 nodes and 249 edges. In this visualisation, nodes are shaded based on their number of connections, effectively highlighting central hubs of interaction within the criminal network.\\}
        \label{fig:meeting_network}
    \end{minipage}
    \begin{minipage}{0.45\textwidth}
        \centering
        \includegraphics[width=\textwidth]{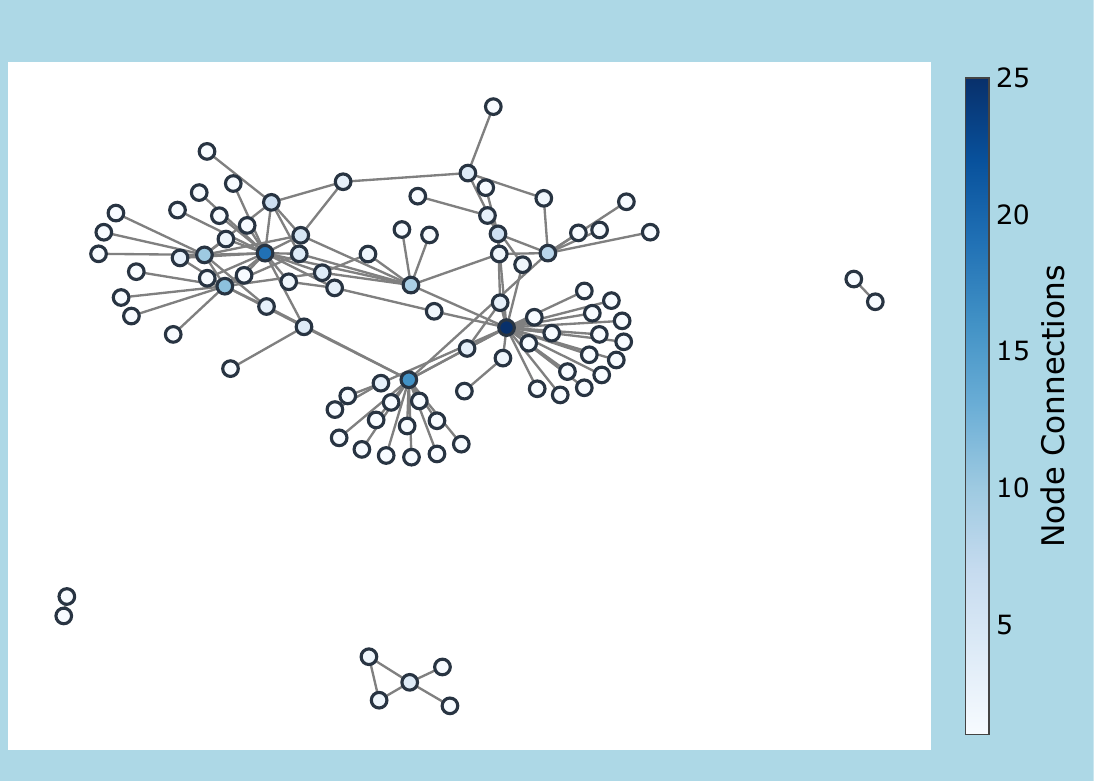}
        \caption{A network constructed using the \textit{phone call} data in the Sicilian Mafia ``Montagna Operation'' dataset, consisting of 94 nodes and 120 edges. Node color intensity indicates the number of connections, and the network structure reveals three nodes that are dominant forming clusters.}
        \label{fig:phone_calls}
    \end{minipage}
\end{figure}

The Sicilian Mafia ``Montagna Operation'' networks are modeled as undirected and weighted graphs due to the reciprocal nature of interactions. Nodes represent individuals, while edges denote the interactions through either physical meetings or phone communications. The undirected edges reflect mutual relationships, while edge weights correspond to the frequency of interactions~\cite{jia2024network}. The dataset includes two distinct networks: the meeting network, which records in-person meetings shown by Fig. \ref{fig:meeting_network}. and the phone call network, which captures telecommunication interactions represented by Fig. \ref{fig:phone_calls}.

\subsection{Competing Approach}

The work in \cite{cavallaro2020disrupting} applied SNA to the ``Montagna Operation'' dataset to examine the structure of Sicilian Mafia. That study explores disruption strategies that simulate real-world police raids and identifies key actors using centrality metrics. Four centrality measures are used:
\begin{itemize}
    \item \emph{Degree centrality} measures a node's influence based on the number of direct connections \cite{ficara2023human}.
    \item \emph{Betweenness centrality} captures how often a node lies on the shortest paths (geodesic) between others, indicating its role as a communication bridge \cite{ficara2023human}.
    \item \emph{Katz centrality} accounts for both immediate and indirect connections to assess a node’s broader influence \cite{cavallaro2020disrupting}.
    \item \emph{Collective influence} evaluates a node’s significance through both direct and indirect links within a defined network radius \cite{cavallaro2020disrupting}.
\end{itemize}
That study also employs weighted versions of these metrics to reflect the frequency of interactions between individuals.

The effectiveness of disruption strategies is assessed using three key metrics:
\begin{itemize}
    \item \emph{Clustering Coefficient (CC)} measures the extent to which nodes form tightly knit groups within the network~\cite{ficara2023human}.
    \item \emph{Average Path Length (APL)} calculates the average shortest (geodesic) path between all pairs of nodes \cite{duxbury2019criminal}.
    \item \emph{Normalised Largest Connected Components} $(\rho)$ represents the proportion of nodes remaining in the largest connected component (LCC) after intervention \cite{cavallaro2020disrupting}.
\end{itemize}

Despite its effectiveness, the work in \cite{cavallaro2020disrupting} presents weaknesses and limitations. It primarily focuses on two objectives, minimising the number of node removals and reducing $\rho$ to measure network fragmentation. This makes the method somewhat rigid, as real-world LEA operations involve dynamic and often conflicting objectives. For example, operational costs as budget, limited personnel, and time influence how many individuals can be targeted and how frequently interventions can occur. That framework does not explicitly account for such operational costs, making it challenging to simultaneously optimise disruption impact and resource efficiency. This represents a significant limitation in applying the method to practical law enforcement scenarios.

\subsection{Proposed Approaches}
The competing methodology proposed by \cite{cavallaro2020disrupting} makes valuable contributions to criminal network disruption through social network analysis (SNA). However, it exhibits key limitations that this paper aims to address:
\begin{itemize}
    \item A narrow objective focus namely only the number of node removals and the normalised largest connected component, making the approach very rigid.
    \item An over-reliance on centrality measures for identifying key players, which may result in suboptimal disruption outcomes.
\end{itemize}

To address the limitation of the competing approach, this paper proposes a multi-objective optimisation framework using the Weighted Sum Genetic Algorithm (WS-GA) and the Non-dominated Sorting Genetic Algorithm II (NSGA-II). The framework simultaneously optimises multiple objectives, including the number of node removal, normalised largest connected component ($\rho$), and spatial distance $D$ as key metrics to reflect the operational constraints faced by LEAs. Additionally, this approach avoids over-reliance on centrality metrics, increasing the likelihood of identifying strategically important, non-central actors. Furthermore, the framework is extensible, allowing additional objectives to be incorporated without major structural changes, making it a scalable and adaptable solution for evolving disruption priorities.

\subsubsection{The Weighted Sum Genetic Algorithm (WS-GA)}
The \emph{WS-GA} is one of the simplest extensions of the single-objective GA for handling multi-objective optimisation problems. It combines multiple objective functions $f_i(\vec{y})$ into a single aggregated function by assigning fixed weights $w_i$ prior to evaluation, allowing the GA to optimise a weighted trade-off among objectives. The aggregated fitness function is generally formulated as:
\begin{equation}
    F(\Vec{y}) = \sum_{i = 1}^{n} w_{i} f_{i}(\Vec{y})
    \label{eq:weighted_sum_apprach}
\end{equation}
where $\Vec{y}$ is the solution vector and the $\sum_{i=1}^{n} w_{i} = 1$ with addition that the weight cannot be changed throughout the iterations. Once the objectives are combined, a standard GA can be applied to optimse $F(\Vec{y})$~\cite{coello2000updated, ismail2011self}.

The weighted sum approach has notable limitations that reduce its effectiveness in complex scenarios. It does not naturally preserve solution diversity, making it prone to premature convergence in local minima. Additionally, selecting appropriate weight values for each objective is challenging and often relies on trial and error \cite{coello2000updated, ismail2011self}.

\subsubsection{Non-Dominated Sorting GA II (NSGA-II)}
The NSGA-II is a widely used evolutionary algorithm in multi-objective optimisation problems. Proposed by \cite{deb2002fast}, it was developed to address the limitations of NSGA. NSGA-II is particularly concerned with computational complexity, elitism, and the need for diversity-preserving parameters. It introduces a fast non-dominated sorting method, a crowding distance calculation selection operator, and elitism by maintaining an archive of the best solutions~\cite{deb2002fast}.

NSGA-II is based on the concept of Pareto dominance, which describes a situation where one solution is better or equal to another in all objectives and strictly better in at least one \cite{YANG2014197}. A solution is considered Pareto optimal if no other solution in the population dominates it, and the collection of such solutions forms the Pareto front.

Then, NSGA-II employs non-dominated sorting to partition the population into multiple front based on Pareto dominance. The first front, $F_1$, contains solutions like $s_1$ for which the number of individuals dominating them, denoted as $n_{s_1}$, is zero \cite{XU201754}. Subsequent fronts are constructed by incrementally including solutions dominated by those in earlier fronts. This sorting mechanism reduces the computational complexity to $O(MN^2)$, where $M$ is the number of objectives and $N$ is the population size \cite{deb2002fast}.

To preserve diversity among solutions within each Pareto front, NSGA-II employs crowding distance calculation. This method estimates the density of solutions by averaging the distance between neighboring solutions in the objective space (within the same front). Boundary solutions are assigned an infinite crowding distance to ensure their retention. Solutions with larger crowding distances are preferred, as they indicate sparser regions in the front, promoting an even distribution of solutions without the need for manually tuned sharing parameters~\cite{deb2002fast, XU201754}. The crowding distance is calculated as follows:
\begin{equation}
    d_s = \sum_{i=1}^{l} \frac{f_{i}(S+1) - f_{i}(S-1)}{f_{i}^{max} - f_{i}^{min}}
\end{equation}
where $d_{s}$ represents the crowding distance assigned to individual $s$, $f_{i}(S+1)$ and $f_{i}(S-1)$ are the objective values of the neighboring solutions, the difference is then normalised by the $f_{i}^{max} - f_{i}^{min}$, which is the range of the $i$-th objective, and lastly $l$ is the number of objectives \cite{deb2002fast, XU201754}.

The selection process in NSGA-II combines elitism and traditional GA operations. It combines the parent and offspring populations and selects the top individuals based on their non-domination rank and crowding distance.  Selection is performed using binary tournament selection with a crowded-comparison operator that favours solutions from lower-ranked Pareto fronts (more optimal) and less crowded regions (more diverse). This elitist strategy preserves high-quality solutions across generations, significantly enhancing convergence toward the Pareto front \cite{deb2002fast, XU201754}. GA operators such as crossover and mutation are then applied to generate new offspring and maintain population diversity \cite{YANG2014197}.

A common characteristic for all multi-objective GAs is the presence of conflicting objectives. This conflict is not a drawback but rather a defining feature that allows algorithms like NSGA-II and WS-GA to explore a diverse set of trade-off solutions. This contradiction is the main reason there is no single optimal solution that satisfies all goals simultaneously, and this gives rise to the Pareto front, where optimal trade-offs can be explored. Without such contradictions, the Pareto front would collapse into a single point and reducing the problem into a single-objective task~\cite{YANG2014197}.

\subsubsection{Objectives}
This section outlines the objectives implemented in the proposed approach. The objectives were selected to allow a direct one-to-one comparison with the competing method, with the addition of a third objective to capture operational costs. Specifically, the model focuses on three primary goals, each optimised using distinct strategies.

\emph{Number of node removal} represents the number of individuals that need to be removed to disrupt the network. This objective is not directly optimised as a fitness value but is instead treated as a parameter sweep, consistent with the competing methodology. This translates to fixed number of nodes removed in each iteration, ranging from 1 to 90. For each fixed value, both WS-GA and NSGA-II are run to optimise the remaining objectives. This enables the identification of the most effective set of individuals to target at each intervention level.

\emph{Normalised largest connected components} ($\rho$) reflects the proportion of nodes remaining in the largest connected component (LCC) of the network after the node removals. In the context of criminal networks, the LCC signifies the most cohesive and structurally resilient segment. Thus, the reduction of LCC translates to network disruption, potential limiting the communication and coordination among the remaining actors \cite{cavallaro2020disrupting}. The metric is computed as follows:
\begin{equation}
    \rho_{i} = 1 - |\frac{LCC(G_{i}) - LCC(G_0)}{LCC(G_0)}|
\end{equation}
where $i$ represents the number of node removals. Thus, $LCC(G_0)$ is the initial LCC of the graph. Lower $\rho$ values indicate higher levels of disruption \cite{cavallaro2020disrupting}. 

It is important to note that $\rho$ is a simplified disruption metric, as it depends heavily on the distribution of connected components. While it works well for the current dataset, where networks are dense and actively connected, it may not generalise to real-world criminal networks, which are often sparse, decentralised, and composed of loosely connected sub-groups \cite{krebs2002uncloaking, krebs2002xmapping}. In such cases, $\rho$ may fail to accurately capture the true extent of disruption.

A key advantage of incorporating $\rho$ instead of the LCC as an objective is its scale-independence. As a normalised metric ranging from zero to one, $\rho$ prevents the optimisation process from becoming biased toward this objective due to disproportionate numerical magnitudes among objectives. However, to align it with the GA framework, a minor adjustment to the equation is required:
\begin{align}
    f_{rho}(\rho_{i}) &= 1 - \rho_{i} \\
    &= |\frac{LCC(G_{i}) - LCC(G_0)}{LCC(G_0)}|
\end{align}
where $f_{rho}(\rho_i)$ is the fitness function that maximises its values so that $\rho$ is minimised.

\emph{Spatial distance} ($D$) quantifies the geographical cost associated with arresting individuals in a criminal network, assuming actors are dispersed across different locations. Specifically, it measures the Euclidean distance between each removed node and the nearest LEA headquarters (LEA HQ). This metric serves as a representative for real-world operational constraints, such as travel time, personnel limitations, and budgetary restrictions. The underlying assumption is that the farther a node is from an LEA HQ, the more difficult and costly it is to reach and apprehend.

To compute this metric, both the network nodes and a predefined number of LEA HQs are randomly assigned synthetic coordinates within a bounded two-dimensional space. The Euclidean distance between each removed node and its nearest HQ is then calculated. These values are then normalised to ensure comparability with other objectives. The calculation is as below:
\begin{equation}
    D_{i} = \frac{1}{|R_i|}(\sum_{k \in R_i} \frac{d_k - d_{min}}{d_{max} - d_{min}})
\end{equation}
where $D_{i}$ is the spatial distance with $i$ being the number of nodes removed. The $R_i$ is a set of removed nodes and $|R_i|$ is the cardinality of subset $R_i$. While $d_k$ is the shortest Euclidean distance of node $k$ to the nearest HQ, and $d_{max}$ and $d_{min}$ are the maximum and minimum Euclidean distance a node could have in the graph. Lower $D_i$ values are preferred, because it indicates lower geographical cost for the LEA. 

To convert this into a maximisation objective for the GA, the following adjustment is applied:
\begin{align}
    f_{spatial}(D_i) &= 1 - D_i \\
    &= 1 - (\frac{1}{|R_i|}(\sum_{k \in R_i} \frac{d_k - d_{min}}{d_{max} - d_{min}}))
\end{align}
where $f_{spatial}(D_i)$ is the fitness function that rewards solutions with lower spatial distance.

The three objectives in this study are interrelated, increasing the number of nodes removed tends to lower $\rho$, increase the level of disruption, but also raises the spatial cost $D$. This reveals a fundamental conflict between disruption effectiveness and operational cost. While maximising disruption is desirable, it must be balanced against the need to minimise spatial costs, highlighting the importance of carefully managing these trade-offs.

Treating the number of node removals as a sweep parameter proves useful in exploring trade-offs more effectively. Including it as an explicit objective could introduce redundancy, as it tends to correlate with either $\rho$ or $D$, depending on the optimisation direction. By instead sweeping across fixed node removal counts, the model gains an additional analytical layer for examining the trade-off between $\rho$ and $D$ metrics.

\subsubsection{Simulations}
This section outlines and justifies the experimental design choices made to generate the results. Simulations involved both meeting and phone-calls dataset from the ``Montagna Operation''. A new attribute called \emph{distance} is created for the networks to calculate the \emph{spatial distance} objective. 

The next step involves replicating the experiment conducted by \cite{cavallaro2020disrupting}, available on \href{https://github.com/lcucav/criminal-nets/tree/master/disruption}{GitHub}. However, this study introduces an additional metric which is the spatial distance into the simulation. This enhancement allows for a more comprehensive comparison with the multi-objective GA later on.

Next, we simulate the proposed approaches using both WS-GA and NSGA-II. The simulation process consists of the following steps:
\begin{enumerate}
    \item Initialisation:

    For each specified number of nodes to remove, both models are initialised using the PyGAD library in Python. Each individual solution in the population represents a unique set of nodes selected for removal from the original graph $G_0$.
    
    \item Disruption and Fitness Evaluation:

    Each solution removes its corresponding nodes, producing a disrupted graph $G_i$. The fitness values are then calculated based on two objectives: the normalised largest connected component ($\rho$) and the spatial distance ($D$).

    \item Objective Handling Based on GA Variant:
    \begin{itemize}
        \item In the WS-GA, the objectives are aggregated into a single fitness value using predefined weights.
        \item In NSGA-II, the objectives are treated as a vector and evaluated using Pareto-based ranking.
    \end{itemize}
    
    \item Iteration:
    
    Steps 2 and 3 are repeated for a defined number of generations within PyGAD, and for each value of node removal (from one to the maximum).
    
\end{enumerate}
The final outcomes of the simulation include the number of nodes removed, the best solution identified for each removal level, and the corresponding values of the normalised largest connected component ($\rho$) and spatial distance ($D$).

One of the most critical factors influencing the efficiency of GAs is the size of the search space, which is the set of all possible solutions the algorithm can explore. In this simulation, the search space is directly affected by how the graph problem is encoded. A common method is binary encoding, where each individual is represented as a binary string of length $|V|$, with 1s indicating node removals. However, this leads to an impractically large search space of size $2^{|V|}$. To address this, we use value encoding, where each solution is a set of node indices corresponding to exactly $i$ nodes to be removed. This approach reduces the search space to ${{|V|} \choose {i}}$, significantly improving computational efficiency.

In GAs, one of the fundamental issues is to balance the trade-off between the exploration, ability to search other potential solutions and exploitation, ability to refine high-quality solution \cite{hussain2020trade}. This balance is crucial for avoiding premature convergence to local minima and increasing the likelihood of identifying a global optimum.

\begin{table}[ht]
    \centering
    \footnotesize
    \caption{The hyperparameters for both NSGA-II and the WS-GA}    
    \begin{tabular}{|p{3cm}|p{1.3cm}|p{2.8cm}|}
    \hline
    \textbf{Parameter} & \textbf{Values} & \textbf{Model}\\ 
    \hline
        Number of Generations & 500 & NSGA-II and WS-GA
        \\
    \hline
        Population Size & 500 & NSGA-II and WS-GA
        \\
    \hline
        Number of Parents & 250 & NSGA-II and WS-GA
        \\
    \hline
        Selection Method & Tournament & NSGA-II and WS-GA
        \\
    \hline
        Crossover Method & Two-point method & NSGA-II and WS-GA
        \\
    \hline
        Mutation Method & Scramble & NSGA-II and WS-GA
        \\
    \hline
        Weight of Each Objective & 0.5 & WS-GA
        \\
    \hline
    \end{tabular}
    \label{tab:MOGA parameters}
\end{table}

Table \ref{tab:MOGA parameters} outlines the parameters used in the implementation of the proposed approaches. In the NSGA-II configuration, tournament selection is employed alongside a two-point crossover method to effectively recombine genetic material from high-quality parents, thereby enhancing exploitation. To promote diversity and support exploration, scramble mutation is applied, helping the algorithm avoid local minima. As NSGA-II evaluates each objective independently, this setup is well-suited for exploring trade-offs and maintaining a diverse set of non-dominated solutions. In contrast, the WS-GA aggregates the $\rho$ and $D$ objectives into a single fitness score using equal weights (0.5 each), ensuring neither metric dominates the optimisation process. Although both algorithms share the same core parameters for consistency, WS-GA reduces the multi-objective problem to a single-objective task, leading the evolutionary process to converge on a single optimal solution rather than a set of trade-offs along the Pareto front.

\section{Results and Discussions}

\subsection{Results}
The outcomes of the simulation outlined in the previous section are organised into two parts. First, a general overview using the Pareto front, highlighting the differences between the competing approach (centrality-based) and the proposed methods (GA-based). This is followed by a deeper analysis of the solutions of the proposed approaches.

\subsubsection{Comparison Between Proposed Approaches and the Competing Approach}
A total of 90 simulations were conducted for both WS-GA and NSGA-II. Each run corresponds to the number of nodes we want to remove. The first run considers the removal of a single node, the second involves two nodes, and this process continues incrementally up to the 90th run, which involves the removal of 90 nodes.

\begin{figure}[ht]
    \centering
    \begin{minipage}{0.42\textwidth}
        \centering
        \includegraphics[width=\textwidth]{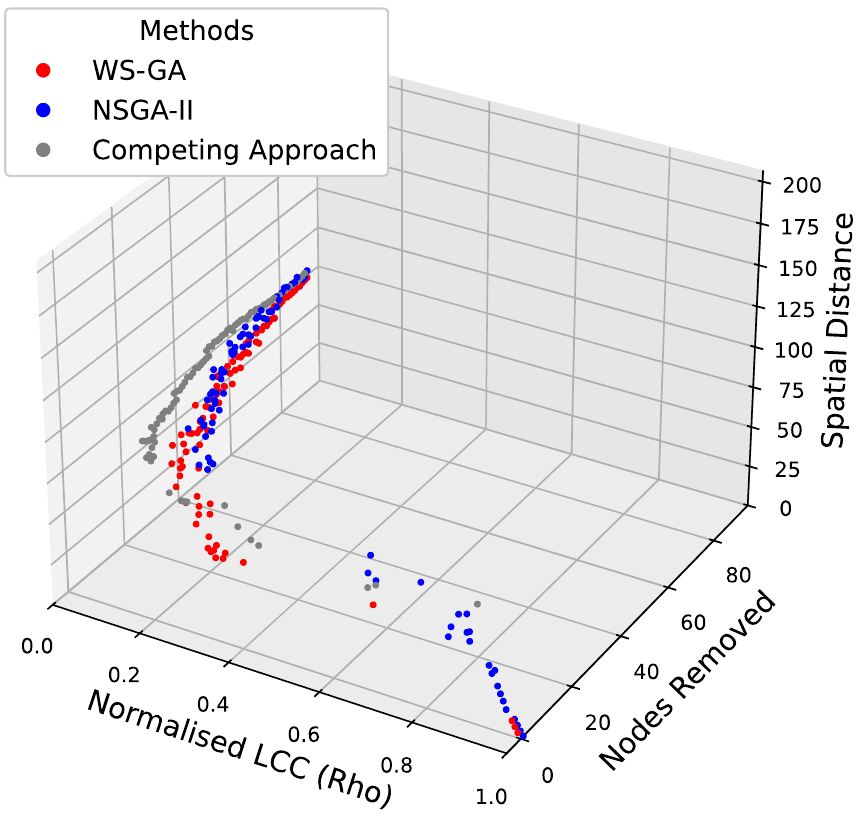} 
        \caption{Combined scattered Pareto front for the meeting dataset, illustrating the trade-off between the Number of Nodes Removed, normalised connected component, and the spatial distance. Each point represents a solution from each strategy.\\}
        \label{fig:pareto_meeting}
    \end{minipage}\hfill
    \begin{minipage}{0.42\textwidth}
        \centering
        \includegraphics[width=\textwidth]{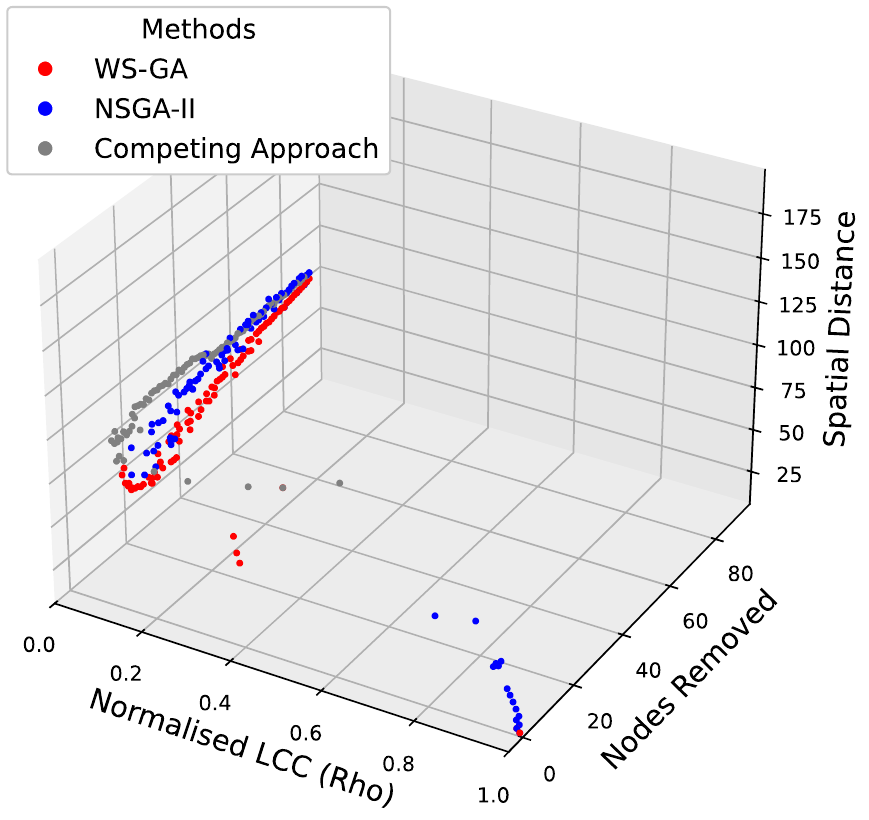}
        \caption{Combined scattered Pareto front for the phone-calls dataset, showing the trade-off between the Number of Nodes Removed, normalised connected component, and spatial distance. Each point represents a solution generated by a different strategy.}
        \label{fig:pareto_phone}
    \end{minipage}
\end{figure}

To evaluate the effectiveness of the proposed approaches against the competing approach, the results are visualised using Pareto front plots in Fig. \ref{fig:pareto_meeting} and Fig. \ref{fig:pareto_phone}. These plots illustrate the trade-offs between the normalised largest connected component ($\rho$), spatial distance ($D$), and the number of nodes removed. A lower $\rho$ value indicates greater fragmentation of the network, signifying more effective disruption. Similarly, a lower $D$ value reflects reduced operational costs.

Each point in Fig. \ref{fig:pareto_meeting} and Fig. \ref{fig:pareto_phone} represents a non-dominated solution for each approach. The points from the competing approach reflect the best-performing results across various centrality and disruption-based strategies, including degree centrality, betweenness centrality, Katz centrality, and collective influence, under both sequential and block removal methods. These optimal points serve as benchmarks for evaluating the effectiveness of the proposed approaches.

Both Fig. \ref{fig:pareto_meeting} and Fig. \ref{fig:pareto_phone} reveal consistent insights regarding the performance of each approach with respect to $\rho$. In the early stages of node removal, the competing approach consistently outperforms both WS-GA and NSGA-II in achieving lower $\rho$ values. However, as the number of node removals increases, the performance of both proposed methods becomes comparable to that of the competing approach, with clear convergence toward similar levels of network fragmentation.

Overall, the proposed approaches demonstrate superior performance in optimising spatial distance ($D$), as evidenced by the colour gradients of the points in the plots. Unlike the competing approach, both WS-GA and NSGA-II explicitly minimise $D$, resulting in significantly lower spatial costs during the first 40 node removals. However, as the number of nodes removed increases, all approaches converge toward similar $D$ values.

\subsubsection{Proposed Approach Analysis}
\begin{figure}[ht]
    \centering
    \includegraphics[width=0.5\textwidth]{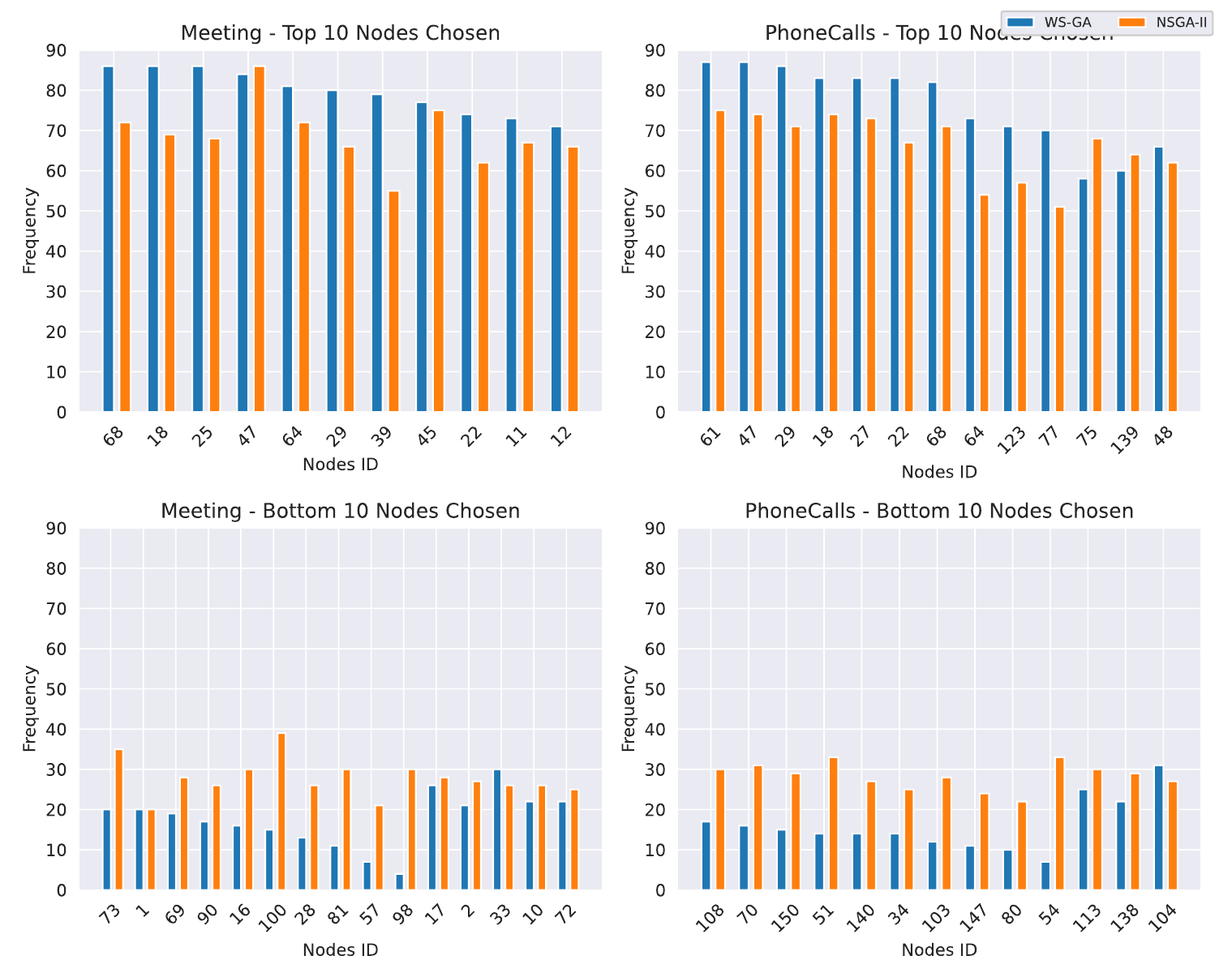}
    \caption{Frequency histograms of chosen nodes in Pareto-optimal solutions for WS-GA and NSGA-II from the ``Montagna Operation`` network. The y-axis shows the frequency of each node being selected in the 90 GA runs across the number of node removals. The top row displays nodes from the union of the top 10 most frequently selected, while the bottom row shows nodes from the union of the bottom 10 least frequently selected nodes by both algorithms. The left column shows the histograms constructed using the meeting dataset and the right column corresponds to phone-calls dataset.}
    \label{fig:selected node hist}
\end{figure}

A closer examination of the nodes selected by WS-GA and NSGA-II is presented in Fig. \ref{fig:selected node hist}. The histograms illustrate the most and least frequently selected nodes across all runs, providing insight into the selection patterns and decision-making criteria of the algorithms. This analysis offers insight to the decision making criteria for each algorithm.
 
Fig. \ref{fig:selected node hist} displays the frequency of nodes selected from the top/bottom 10 nodes for both WS-GA and NSGA-II across the meeting and the phone-calls datasets. The top row of the figure shows that both algorithms tend to prioritise a similar subset of nodes, as reflected by the limited number of distinct nodes on the x-axis. A similar pattern is observed in the bottom row of the figure, reinforcing the conclusion that both WS-GA and NSGA-II consistently target a common set of nodes.

The figure also reveals that WS-GA selects certain nodes with a frequency close to 90, indicating that these nodes are consistently prioritised across nearly all runs. This suggests that the algorithm identifies them as highly influential early targets. Conversely, some nodes appear with low frequency in the lower part of the figure, implying that WS-GA does not prioritise these nodes and selects them only in the later stages of number of nodes removed.

\begin{table}[ht]
    \centering
    \small
    \caption{Top Five Most Frequent Overlapping Nodes in WS-GA and NSGA-II in ``Montagna Operation''}
    \begin{tabular}{|p{0.8cm}|p{2.9cm}|p{1.5cm}|p{1.6cm}|}
    \hline
    \textbf{Nodes} & \textbf{Role} & \textbf{Degree Centrality} & \textbf{Betweenness Centrality}\\ 
    \hline
        47 & Leader (Boss) & 0.209 & 0.193
        \\
    \hline
        18 & Executive Family & 0.257 & 0.356
        \\
    \hline
        68 & Executive Family & 0.165 & 0.149
        \\
    \hline
        29 & Entrepreneur & 0.118 & 0.155
        \\
    \hline
        22 & Pharmacist (Member) & 0.112 & 0.080
        \\
    \hline
    \end{tabular}
    \label{tab:top5_common nodes}
\end{table}

Table \ref{tab:top5_common nodes} presents the top five most frequently selected nodes by both WS-GA and NSGA-II, based on the combined selection frequency across the ``Montagna Operation'' networks. The table includes each node's associated role and average centrality measures, calculated from the original, undisrupted network. Notably, node 47 identified as a Leader/Boss, which is one of the most frequently selected nodes, supporting previous findings. This node exhibits the second-highest degree and betweenness centrality, highlighting its strategic importance. Other frequently selected nodes, such as 18 and 68, also demonstrate notably high centrality values. Overall, the table reinforces the correlation between high centrality measures and selection frequency by both GA-based methods.

\begin{table}[ht]
    \centering
    \footnotesize
    \caption{Performance comparison of LEA's real-world arrests with competing methods}
    \begin{tabular}{|p{2cm}|p{1.2cm}|p{1cm}|p{1cm}|p{1.3cm}|}
    \hline
    \textbf{Metric} & \textbf{Real-World} & \textbf{WS-GA} & \textbf{NSGA-II} & \textbf{Centrality Baseline} \\ 
    \hline
        Number of Nodes Removed & 41 & 41 & 41 & 41
        \\
    \hline
        Normalised LCC ($\rho$) & 0.093 & 0.053 & 0.052 & 0.041
        \\
    \hline
        Spatial Distance ($D$) & 76.653 & 57.258 & 74.348 & 81.538
        \\
    \hline
    \end{tabular}
    \label{tab:comparison arrested vs model}
\end{table}

Table \ref{tab:comparison arrested vs model} evaluates the effectiveness of the proposed approaches against both the real-world LEA arrests and the competing approach (i.e., centrality baseline). The competing approach applies sequential unweighted betweenness centrality for node removal \cite{cavallaro2020disrupting}. The table compares $\rho$ and $D$ at a fixed number of nodes removed. The values reported for WS-GA, NSGA-II, and the competing approach represent the average performance across both the meeting and phone-calls datasets, ensuring a balanced comparison. This comparison offers insight into how closely the optimised disruption strategies align with real-world LEA arrests.

The results in Table \ref{tab:comparison arrested vs model} indicate that, in terms of network disruption ($\rho$), the proposed approaches and the competing method outperform the real-world LEA arrests. Among them, the competing approach achieves the lowest $\rho$, demonstrating the most effective fragmentation of the network. WS-GA and NSGA-II produce $\rho$ values that are relatively close, indicating competitive performance. In terms of spatial distance ($D$), WS-GA clearly outperforms all other methods, followed by NSGA-II. In contrast, the competing approach performs the worst in $D$, as it does not incorporate spatial cost into its optimisation process.

\subsection{Discussions}
First, both WS-GA and NSGA-II consistently underperform compared to the competing approach in terms of $\rho$ when the number of nodes removed is small. However, as the number of nodes removed increases, the structural disruption achieved by all approaches is comparable. 

Another key factor is the trade-off between $\rho$ and $D$. While the proposed approaches consistently outperform the competing method in terms of $D$, this often comes at the cost of slightly higher $\rho$. This is expected because WS-GA and NSGA-II optimise both objectives, whereas the competing approach does not account for $D$. As more nodes are removed, both $\rho$ and $D$ values converge across all methods, indicating increasingly similar node selections and diminishing returns from further optimisation.

Notably, NSGA-II performs poorly on $\rho$ but well on $D$ when the number of node removals is below 30. This occurs because NSGA-II treats the objectives separately, unlike WS-GA, which combines them into a single weighted objective. Thus, NSGA-II tends to prioritise diversity and elitism through Pareto dominance and crowding distance. In the early stages of the number of nodes removed, $\rho$ values change very little, while $D$ fluctuates more significantly. This leads NSGA-II to favour solutions optimising $D$, causing it to remain in the exploratory phase longer. Consequently, NSGA-II often requires more generations and a larger population size to reach competitive performance. In contrast, WS-GA converges more quickly but is more prone to getting stuck in local minima.

Apart from the differences on how the proposed approaches work, they are frequently choosing the same set of nodes, suggesting a shared understanding of which nodes are critical to the objectives. This also implies that both WS-GA and NSGA-II follow a similar hierarchy of node importance. Furthermore, the proposed approaches frequently select nodes with high centrality values and key roles. This demonstrates that these methods naturally align with the competing approach and the real-world scenarios. 

The proposed approaches outperform both the LEA’s real-world arrest strategy and the competing method, with only a slight shortfall in $\rho$. Between the two, WS-GA performs better overall due to its faster convergence and stronger exploitation of high-quality solutions. In contrast, NSGA-II requires more time to explore the solution space and, under limited generations, struggles to reach optimal performance. However, as highlighted in the results, NSGA-II has the potential to surpass WS-GA in the long term, particularly due to its ability to preserve diversity and avoid local minima. This suggests that for more complex or harder optimisation objectives, NSGA-II may offer greater potential.

\section{Conclusion and Future Work}
This study proposed the Weighted Sum Genetic Algorithm (WS-GA) and the Non-dominated Sorting Genetic Algorithm II (NSGA-II) as multi-objective optimisation methods for disrupting criminal networks. These algorithms are designed to balance the trade-off between structural fragmentation and operational cost management, represented by the normalised connected component ($\rho$) and spatial distance ($D$), respectively. This approach extends and improves upon the competing method introduced by \cite{cavallaro2020disrupting}. While the competing approach, grounded in classical centrality metrics, performs better in early-stage fragmentation, both WS-GA and NSGA-II eventually converge and deliver comparable performance in $\rho$. More importantly, the proposed methods significantly outperform the competing approach in spatial distance, with WS-GA consistently offering the best trade-off across objectives. Additionally, the proposed approaches can identify key players without relying on centrality metrics and roles, allowing them to detect influential peripheral nodes that might otherwise be overlooked.

Despite these promising results, several limitations must be acknowledged. Both proposed algorithms operate under the assumption of a static network. In practice, however, criminal networks are dynamic as they adapt, reorganise, and respond to disruption efforts. This static assumption limits the applicability of the current results to real-world scenarios. Future models should incorporate methods for handling temporal evolution and adaptive behaviours such as deep reinforcement learning~\cite{nguyen2019multi} to more accurately capture the complexities of criminal networks and to develop disruption strategies that remain effective over time.

Furthermore, computational complexity remains a key concern, particularly due to the multi-objective nature of the GA models. This is especially evident in NSGA-II, which demands a large population and many generations to explore and converge effectively. Although WS-GA is relatively faster, it still incurs higher runtime than the competing approach. Future work could explore acceleration strategies~\cite{cheng2019accelerating} to enhance algorithmic efficiency and reduce runtime.

\bibliographystyle{IEEEtran}
\bibliography{cas-refs}

@article{cheng2019accelerating,
  title={Accelerating genetic algorithms with {GPU} computing: A selective overview},
  author={Cheng, John Runwei and Gen, Mitsuo},
  journal={Computers \& Industrial Engineering},
  volume={128},
  pages={514--525},
  year={2019},
  publisher={Elsevier}
}

@article{cavallaro2020disrupting,
  title={Disrupting resilient criminal networks through data analysis: The case of {Sicilian Mafia}},
  author={Cavallaro, Lucia and Ficara, Annamaria and De Meo, Pasquale and Fiumara, Giacomo and Catanese, Salvatore and Bagdasar, Ovidiu and Song, Wei and Liotta, Antonio},
  journal={PLoS One},
  volume={15},
  number={8},
  pages={e0236476},
  year={2020},
  publisher={Public Library of Science San Francisco, CA USA}
}

@inproceedings{ficara2021multilayer,
  title={Multilayer network analysis: the identification of key actors in a {Sicilian Mafia} operation},
  author={Ficara, Annamaria and Fiumara, Giacomo and De Meo, Pasquale and Catanese, Salvatore},
  booktitle={Future Access Enablers of Ubiquitous and Intelligent Infrastructures},
  pages={120--134},
  year={2021},
  organization={Springer}
}

@article{krebs2002uncloaking, title={Uncloaking Terrorist Networks}, volume={7}, url={https://firstmonday.org/ojs/index.php/fm/article/view/941}, DOI={10.5210/fm.v7i4.941}, abstractNote={This paper looks at mapping covert networks using data available from news sources on the World Wide Web. Specifically, we examine the network surrounding the tragic events of September 11th 2001. Through public data we are able to map a portion of the network centered on the 19 dead hijackers. This map gives us some insight into the terrorist organization, yet it is incomplete. Suggestions for further work and research are offered.}, number={4}, journal={First Monday}, author={Krebs, Valdis}, year={2002}, month={Apr.} }

@article{xu2005criminal,
  title={Criminal network analysis and visualization},
  author = {Xu, Jennifer and Chen, Hsinchun},
  journal={Communications of the ACM},
  volume={48},
  number={6},
  pages={100--107},
  year={2005},
  publisher={ACM New York, NY, USA}
}

@article{ficara2023human,
  title={Human and social capital strategies for Mafia network disruption},
  author={Ficara, Annamaria and Curreri, Francesco and Fiumara, Giacomo and De Meo, Pasquale},
  journal={IEEE Transactions on Information Forensics and Security},
  volume={18},
  pages={1926--1936},
  year={2023},
  publisher={IEEE}
}

@article{duxbury2020responsiveness,
  title={The responsiveness of criminal networks to intentional attacks: Disrupting darknet drug trade},
  author={Scott W. Duxbury and Dana L. Haynie},
  journal={PLoS One},
  volume={15},
  number={9},
  pages={e0238019},
  year={2020},
  publisher={Public Library of Science San Francisco, CA USA}
}

@article{nguyen2019multi,
  title={Multi-agent behavioral control system using deep reinforcement learning},
  author={Nguyen, Ngoc Duy and Nguyen, Thanh and Nahavandi, Saeid},
  journal={Neurocomputing},
  volume={359},
  pages={58--68},
  year={2019},
  publisher={Elsevier}
}

@article{jia2024network,
  title={Network disruption via continuous batch removal: The case of {Sicilian Mafia}},
  author={Jia, Mingshan and De Meo, Pasquale and Gabrys, Bogdan and Musial, Katarzyna},
  journal={PLoS One},
  volume={19},
  number={8},
  pages={e0308722},
  year={2024},
  publisher={Public Library of Science San Francisco, CA USA}
}

@article{duxbury2019criminal,
  title={Criminal network security: An agent-based approach to evaluating network resilience},
  author={Scott W. Duxbury and Dana L. Haynie},
  journal={Criminology},
  volume={57},
  number={2},
  pages={314--342},
  year={2019},
  publisher={Wiley Online Library}
}

@article{krebs2002xmapping,
  title={Mapping networks of terrorist cells},
  author={Krebs, Valdis},
  journal={Connections},
  volume={24},
  number={3},
  pages={43--52},
  year={2002}
}

@article{hussain2020trade,
  title={Trade-off between exploration and exploitation with genetic algorithm using a novel selection operator},
  author={Hussain, Abid and Muhammad, Yousaf Shad},
  journal={Complex \& Intelligent Systems},
  volume={6},
  number={1},
  pages={1--14},
  year={2020},
  publisher={Springer}
}

@article{coello2000updated,
  title={An updated survey of {GA}-based multiobjective optimization techniques},
  author={Coello, Carlos A},
  journal={ACM Computing Surveys (CSUR)},
  volume={32},
  number={2},
  pages={109--143},
  year={2000},
  publisher={ACM New York, NY, USA}
}

@article{ismail2011self,
  title={Self organizing multi-objective optimization problem},
  author={Ismail, Fatimah Sham and Yusof, Rubiyah and Khalid, Marzuki},
  journal={International Journal of Innovative Computing, Information and Control},
  volume={7},
  number={1},
  pages={301--314},
  year={2011}
}

@article{deb2002fast,
  title={A fast and elitist multiobjective genetic algorithm: {NSGA-II}},
  author={Deb, Kalyanmoy and Pratap, Amrit and Agarwal, Sameer and Meyarivan, TAMT},
  journal={IEEE Transactions on Evolutionary Computation},
  volume={6},
  number={2},
  pages={182--197},
  year={2002},
  publisher={Ieee}
}

@incollection{YANG2014197,
title = {Chapter 14 - {Multi-Objective Optimization}},
editor = {Xin-She Yang},
booktitle = {Nature-Inspired Optimization Algorithms},
publisher = {Elsevier},
address = {Oxford},
pages = {197-211},
year = {2014},
isbn = {978-0-12-416743-8},
doi = {https://doi.org/10.1016/B978-0-12-416743-8.00014-2},
author = {Xin-She Yang}
}

@article{XU201754,
title = {Multi-objective based spectral unmixing for hyperspectral images},
journal = {ISPRS Journal of Photogrammetry and Remote Sensing},
volume = {124},
pages = {54-69},
year = {2017},
issn = {0924-2716},
doi = {https://doi.org/10.1016/j.isprsjprs.2016.12.010},
author = {Xu, Xia and Shi, Zhenwei},
}

@incollection{mcandrew2021structural,
  title={The structural analysis of criminal networks},
  author={McAndrew, Duncan},
  booktitle={{The Social Psychology of Crime}},
  pages={51--94},
  year={2000},
  publisher={Routledge}
}

@article{ficara2022covert,
  title={Covert network construction, disruption, and resilience: a survey},
  author={Ficara, Annamaria and Curreri, Francesco and Fiumara, Giacomo and De Meo, Pasquale and Liotta, Antonio},
  journal={Mathematics},
  volume={10},
  number={16},
  pages={2929},
  year={2022},
  publisher={MDPI}
}

@incollection{bright2015disrupting,
year = {2015},
language = {eng},
pages = {39-51},
publisher = {Cambridge University Press},
title = {Disrupting and Dismantling Dark Networks: Lessons from Social Network Analysis and Law Enforcement Simulations},
author = {Bright, David A.},
booktitle = {Illuminating Dark Networks},
copyright = {Cambridge University Press 2016},
isbn = {9781107500884},
}

@article{basu2021identifying,
  title={Identifying individuals associated with organized criminal networks: A social network analysis},
  author={Basu, Kaustav and Sen, Arunabha},
  journal={Social Networks},
  volume={64},
  pages={42--54},
  year={2021},
  publisher={Elsevier}
}

\end{document}